\documentclass[letterpaper]{article} %
\usepackage{aaai2026}  %
\usepackage{times}  %
\usepackage{helvet}  %
\usepackage{courier}  %
\usepackage[hyphens]{url}  %
\usepackage{graphicx} %
\usepackage{natbib}  %
\usepackage{caption} %
\usepackage{algorithm}
\usepackage{algorithmic}

\usepackage{newfloat}
\usepackage{listings}

\usepackage{microtype}

\usepackage{amsmath}
\usepackage{amssymb}
\DeclareCaptionStyle{ruled}{labelfont=normalfont,labelsep=colon,strut=off} %
\floatstyle{ruled}
\newfloat{listing}{tb}{lst}{}
\floatname{listing}{Listing}
\newcommand{\possessivecite}[1]{\citeauthor{#1}'s \citeyearpar{#1}}

\title{Beyond World Models: Rethinking Understanding in AI Models}

\author{
    Tarun Gupta, Danish Pruthi
}
\affiliations{
    Indian Institute of Science\\
    Bengaluru, KA, India\\
    \texttt{\{tarungupta, danishp\}@iisc.ac.in}
}

\begin{document}

\maketitle

\begin{abstract}
    World models have garnered substantial interest in the AI community. These are internal representations that simulate aspects of the external world, track entities and states, capture causal relationships, and enable prediction of consequences. This contrasts with representations based solely on statistical correlations. A key motivation behind this research direction is that humans possess such mental world models, and finding evidence of similar representations in AI models might indicate that these models ``understand'' the world in a human-like way. In this paper, we use case studies from the philosophy of science literature to critically examine whether the world model framework adequately characterizes human-level understanding. We focus on specific philosophical analyses where the distinction between world model capabilities and human understanding is most pronounced. While these represent particular views of understanding rather than universal definitions, they help us explore the limits of world models.
\end{abstract}

\section{Introduction}
In artificial intelligence, the concept of world models raises fundamental questions across domains: 
Do LLM representations track world states and the transitions between them, and do they use these representations to predict next tokens?
Do video-generation models create representations of physical laws and spatial geometry, predicting future frames by simulating these learned laws of nature? 
At its core, the world model hypothesis asks whether neural networks capture and reproduce the actual causal processes that generated their data, or whether they merely manipulate surface patterns and capture correlations without intermediate representations that mirror real-world mechanisms \cite{andreas2024worldmodels}. 

The motivation for studying world models stems from the human experience of mental visualization and picturing, along with our ability to mentally simulate these visualized mental models.
A quintessential example is the heliocentric model of the solar system, where humans visualize the sun, planets, and other celestial bodies as entities with specific states (positions, velocities) that transition according to physical laws governing their orbits.\footnote{While one commonly speaks of mental models of the `real world,' these are more precisely models of our theories about the world, such as the heliocentric model itself, or even superseded theories like the geocentric model with its epicycles, or Bohr's model of electrons orbiting the atomic nucleus in discrete paths.}

This intuitive appeal of world models raises the question: If AI models (e.g., LLMs) can maintain such world states and model state transitions rather than just leveraging surface-level correlations, would this constitute human-like understanding? 
It is often argued that since mental world models are an integral component of how humans understand the physical world, the presence of world models in AI models suggests human-like understanding capabilities \cite{lecun2022path, ng2023aiunderstand,mitchell2024worldmodels1,delser2025worldmodelsartificialintelligence}.\footnote{See \S\ref{app:quotes} for specific quotations from these cited works on the connection between world models and understanding.}
While both world models and understanding lack universally agreed-upon definitions, the growing interest in world models within AI research makes this question important to examine. In this paper, we argue that while world models represent a crucial advance beyond mere surface patterns, they fail to capture human-level understanding across various domains of physical reasoning and problem-solving.

To be clear, our argument is not that AI models cannot achieve understanding. We do not claim that current or future AI models lack the capacity to understand the world. Instead, we critique the world model framework as an inadequate theoretical lens for characterizing understanding. AI models may well develop understanding through mechanisms that go beyond or differ entirely from world models. Our critique targets the specific claim that possessing world model-like representations constitutes understanding.

Rather than attempting to provide universal criteria for understanding across all domains, we adopt a case study approach that examines particular, yet fundamental, instances of understanding where the limitations of the world model framework become apparent. Understanding is multifaceted---in any given domain, there are multiple valid perspectives on what constitutes understanding. Our strategy is deliberately selective: we focus on particular philosophical analyses that illustrate aspects of understanding that go beyond what world models can capture. 
We acknowledge that other perspectives on understanding might align better with world models.
However, our selected cases help us explore limitations of using world models to characterize human-level understanding.

To this end, we examine three cases from philosophical work: (1) \possessivecite{hofstadter2007strange} analysis of a computer built from falling dominoes, (2) \possessivecite{Poincare1914-POISAM} distinction between verifying and understanding mathematical proofs, and (3) \possessivecite{popper1979objective} account of understanding physical theories through their problem situations. Through these case studies, we show that the world-model conception fails to capture what these philosophical analyses reveal about human-level understanding. Our analysis contributes to discussions about world model research and its theoretical foundations.

\section{Background and Related Work}
\label{app:related-work}

\paragraph{World Models in AI.} 
\label{app:para:world-models}
The term ``world model'' draws considerable attention in the AI community and is considered a key ingredient for building general intelligence \cite{lecun2022path,ding2024understanding}. 
However, it lacks a universally-accepted definition. Different researchers define world models differently \cite{ding2024understanding, xing2025critiques}. For this paper, we define world models---following the prevailing conception---as internal representations that track objects, their states, and the rules governing how states change over time.

To investigate whether AI models develop such world models, researchers use probing techniques to examine internal representations in neural networks. This involves analyzing learned features in specific layers, studying activation patterns, or using linear decoding methods to recover world state representations from the model's internal states.
These probing approaches aim to determine whether models maintain interpretable representations of world states and transitions rather than relying solely on surface-level pattern matching. For LLMs, \citet{li2023emergent} achieve a landmark result using such probing techniques, showing that a language model trained to play the board game Othello developed an internal world model of the game. Specifically, they find that the model's internal representations can be linearly decoded to recover the actual board state at each move, suggesting the model maintains an internal simulation of game states and transitions rather than relying solely on surface-level pattern matching in move sequences. Similar evidence of internal world models has been found in LLMs trained on chess \cite{karvonen2024emergent}. On the other hand, some research suggests these world models are not clean, human-like mental models, but collections of learned heuristics \cite{lesswrong2024othellogpt,nikankin2024arithmetic}. For an excellent discussion of this topic, we refer readers to \cite{mitchell2024worldmodels2}.

Advances in multimodal models have expanded the use of the term ``world models'' to include video generation systems. Models like Sora \cite{openai2024sora}, WorldGPT \cite{yang2024worldgpt} and Veo \cite{deepmind2025veo} are called world models because they generate videos that appear to follow physical laws and maintain temporal consistency. However, this capability to produce visually plausible sequences is distinct from the question of whether a model's internal representations track discrete states and model transitions between them. 
Likewise, other categories of models, including gaming world models \cite{bruce2024geniegenerativeinteractiveenvironments}, 3D scene models \cite{worldlabs2025generating} and physical-world models \cite{nvidia2025cosmosworldfoundationmodel}, are also sometimes termed world models, each with different definitions and capabilities \cite{xing2025critiques}. Such models are not relevant to our discussion, which examines whether the presence of world model-like representations that explicitly track states and transitions in AI models constitutes human-like understanding.

\paragraph{Philosophical Perspectives on Understanding.} 
\label{app:para:understanding}
``Understanding'' has been a subject of intense debate in philosophy of science and philosophy of mathematics. 
For a general overview of the literature on understanding, see \cite{Grimm2011-GRIU,Grimm2017-STEQAT-2,baumberger2016understanding}. 
For discussions specifically focused on mathematical understanding, see \cite{avigad2008understanding,hamami2024understanding}.
Explanation and understanding are closely related, where the latter is seen as the goal of the former \cite{friedman1974explanation,grimm2010goal}.

Given the numerous theories and frameworks proposed to explain the nature of understanding, arriving at a single, unified account is challenging. Therefore, we avoid defining ``understanding'' in this paper, as providing a definition---even an operational one---would be counter-productive. Understanding remains a contested concept in epistemology without agreed definition, and imposing one would produce only false precision. Any operational definition would merely shift the problem to the defining terms, leading to an infinite regress unless we admit primitive terms, that is, undefined terms \cite{popper1945open}. Rather than impose an arbitrary definition on a concept unsettled in the broader literature, we take an orthogonal approach---using case studies where aspects of understanding are apparent from prior analyses and showing how the world-model conception of understanding falls short of explaining human-level understanding. 

\section{Case Studies: Understanding Beyond World Models}
\label{sec:case-studies}

This section examines three cases where the world model conception falls short of characterizing human-level understanding: a computer built from falling dominoes (\S\ref{subsec:understanding-dominoes}), mathematical proofs (\S\ref{subsec:understanding-proofs}), and Bohr's atomic theory (\S\ref{subsec:understanding-physical-theories}).

\subsection{Understanding a Computer Built from Dominoes}
\label{subsec:understanding-dominoes}

\begin{figure*}[h]
    \centering
    \includegraphics[width=0.9\textwidth]{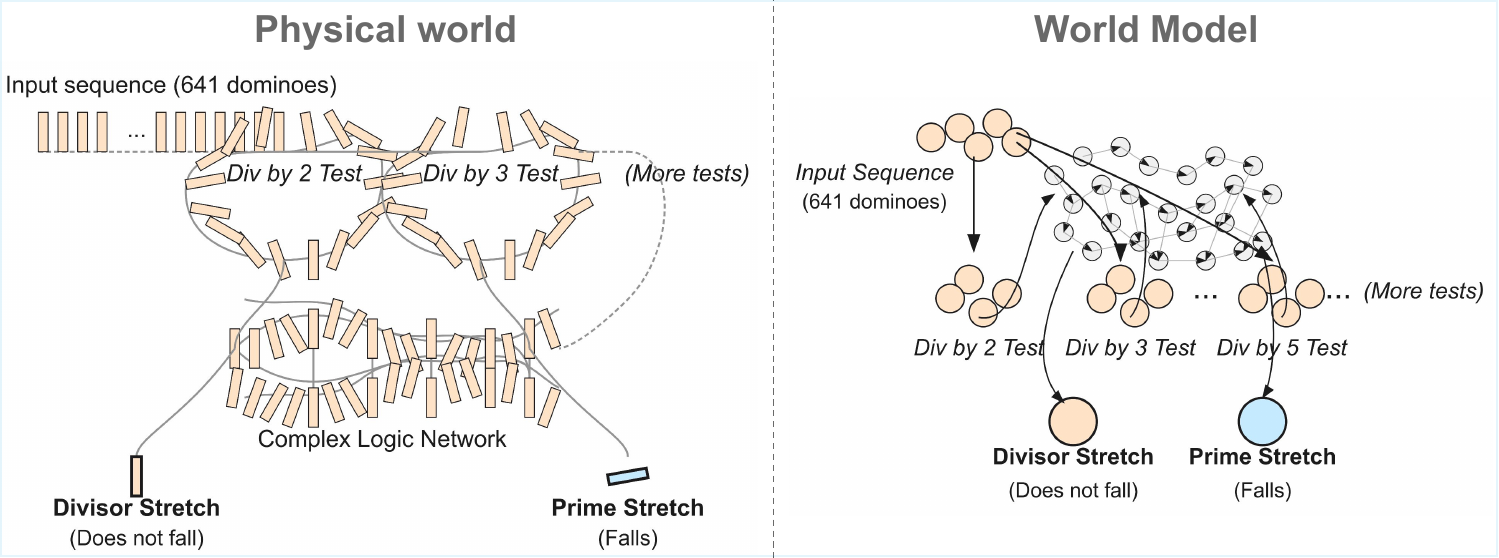}
    \caption{Left: Conceptual illustration of physical arrangement of dominoes in a computational system \cite{hofstadter2007strange}. Right: A schematic world-model representation showing states and causal relationships between dominoes. While the world model can track physical states (standing or fallen dominoes) and predict how one domino causes another to fall, it fails to capture the abstract mathematical concept of primality that fundamentally explains the system's behavior. 
    }
    \label{fig:dominoes}
\end{figure*}

The prevailing conception in world model research holds that understanding physical scenes involves maintaining states for discrete, recognizable objects and tracking the intuitive physics relationships between them. These states, which researchers often probe for in the internal representations of models, correspond to what seems meaningful to humans as states. For example, in board games, each square represents a distinct state rather than the microscopic wood grain patterns or exact molecular arrangements of the board material. Similarly, in video understanding, states might track objects like cars or people rather than individual pixel intensities or spectral frequencies. However, when we apply this approach of state selection to \possessivecite{hofstadter2007strange} thought experiment, we see it can miss understanding a system's behavior.

In this thought experiment, dominoes are arranged into a mechanical computer that determines whether numbers are prime. When a domino falls, it pops back up after a fixed time, thereby propagating signals along carefully-arranged networks. With such a system, we can implement a mechanical computer where signals travel down stretches of dominoes that bifurcate, join together, propagate in loops, and jointly trigger other signals. Relative timing is of course crucial, but the specific implementation details are not relevant to our discussion. The basic idea is that a precisely arranged network of domino chains can function as a computer program for carrying out computations---in this case, determining if a number is prime. (See \S\ref{app:domino} for a detailed description of this mechanical computer.)

To test primality, the system takes input by placing exactly that many dominoes (e.g., $641$) end-to-end in a designated ``input stretch.'' When triggered, the system runs various tests for divisibility by potential factors. If any divisor is found, a signal travels down a specific ``divisor stretch,'' indicating the number is not prime. Conversely, if no divisors are found, a signal travels down a ``prime stretch,'' confirming primality. This arrangement is shown schematically in Figure~\ref{fig:dominoes}.

A world model approach to understanding this system would track each domino's position (standing or fallen) at each moment and simulate the physical propagation of falling patterns. When examining why a particular domino never falls when the input is $641$, such an approach would focus on the immediate physical causes: none of its neighboring dominoes ever fall. This answer, while physically accurate, merely shifts attention to other dominoes. Tracing backward through the causal chain would eventually lead to a statement of the kind: ``That domino did not fall because none of the patterns of motion initiated by the first domino ever include it.'' This mechanistic tracking of states fails to capture understanding of this system's behavior. The key to understanding lies in recognizing that $641$ is prime, an abstract mathematical property that explains the entire pattern of domino behaviors. This understanding cannot be obtained by simply tracking the states of dominoes falling or not falling---no amount of state tracking can reveal the fundamental mathematical concept of primality that governs the system's behavior.

A natural response to this analysis might be that world models could be enriched to include abstract concepts like primality as states themselves---perhaps representing ``641 is prime'' as a state connected to the physical domino configurations. While such an approach would address this critique, it would come at the price of reducing the framework's explanatory power and limiting its falsifiability. We address this counterargument and its broader implications in \S\ref{sec:counterarguments}.

\subsection{Understanding Mathematical Proofs}
\label{subsec:understanding-proofs}

\begin{figure}[t]
    \centering
    \includegraphics[width=0.4\textwidth]{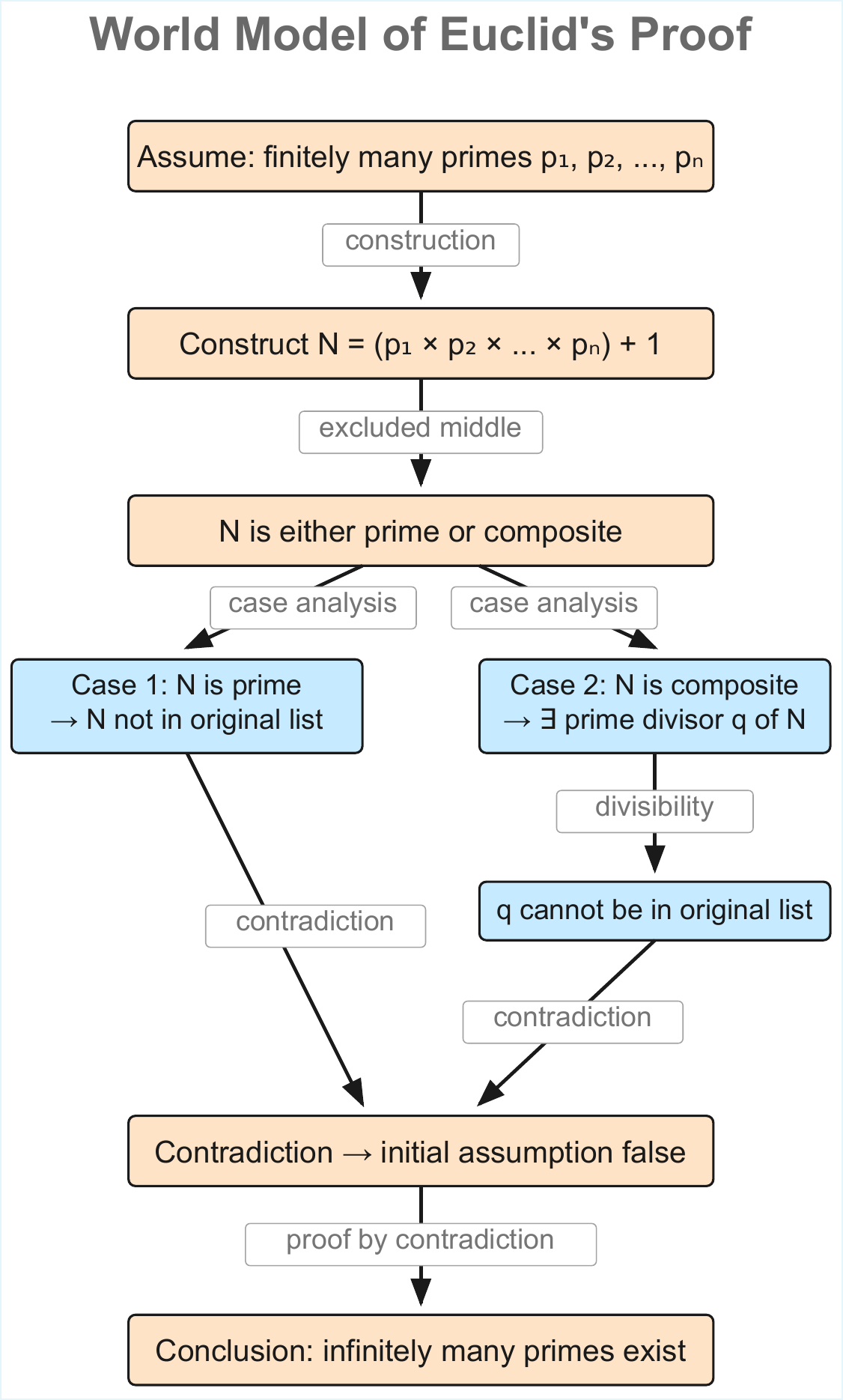}
    \caption{A world model representation of Euclid's proof showing logical states and transitions between them.}
    \label{fig:euclid_world_model}
\end{figure}

In Turing's theory, computations are essentially the same thing as proofs: every valid proof can be converted to a computation that computes the conclusion from the premises, and every correctly executed computation is a proof that the output is the outcome of the given operations on the input. This fundamental equivalence, formalized in the Curry-Howard correspondence \cite{howard1980formulae}, establishes that mathematical proofs are sequences of logical transformations that can be viewed as physical processes: step-by-step manipulations of symbolic expressions according to formal rules. World models, though usually conceived for understanding aspects of physical reality, are equally applicable to proof-understanding given that proofs constitute physical processes of symbolic manipulation. \citet{hu2023language} also argue that world models are important for mathematical reasoning more broadly, suggesting that explicit modeling of intermediate mathematical conclusions and internal simulation of future states is required for mathematical reasoning.
Given this connection between world models and mathematical reasoning, and proof-understanding being a subset of mathematical reasoning with extensive philosophical literature \cite{avigad2008understanding, hamami2024understanding}, proof-understanding offers a focused domain for examining whether world model approaches adequately capture human-level understanding.

A world model approach to proof-understanding would treat proofs as sequences of state transitions, tracking the logical validity of each step. 
When asked why a particular conclusion holds, such a model would trace backward through the chain of logical states. For example, consider Euclid's famous proof that there are infinitely many primes \cite{heath1956thirteen} (see \S\ref{app:euclid} for the proof). Figure~\ref{fig:euclid_world_model} illustrates how a world model would represent this proof as a sequence of logical state transitions. The final state might show ``Therefore, there are infinitely many prime numbers.'' The immediately preceding state might contain ``Since our assumption led to a contradiction, the original assumption that there are finitely many primes must be false.'' 
Working backwards, we might find ``Consider $N+1$, where $N$ is the product of all primes on our list.'' 
This approach amounts to mere verification---confirming that each state transition (logical step) adheres to the rules of logical deduction and that the chain of states connects the premises to the conclusion.
But does such verification constitute human-like understanding? No. It's a common observation in mathematics that there is an important difference between understanding a proof and verifying it. 
As \citet{Poincare1914-POISAM} observes\footnote{This distinction is widely remarked upon, not just by \citet{Poincare1914-POISAM}; for collections and philosophical discussion of many such statements by mathematicians, see \cite{avigad2008understanding,hamami2024understanding}.}:

\begin{quotation}
\textit{Does understanding the demonstration of a theorem consist in examining each of
the syllogisms of which it is composed in succession, and being convinced that it is
correct and conforms to the rules of the game? [...]}

\textit{Yes, for some it is; when they have arrived at the conviction, they will say, I understand. But not for the majority. Almost all are more exacting; they want to know
not only whether all the syllogisms of a demonstration are correct, but why they are
linked together in one order rather than in another. As long as they appear to them
engendered by caprice, and not by an intelligence constantly conscious of the end to
be attained, they do not think they have understood}
\end{quotation}

For an agent with a world-model conception of understanding, the state transitions in a proof appear as if ``engendered by caprice.'' While world models may help in simulating future states of proof steps---as suggested by \cite{hu2023language} to be important for reasoning---this capability alone cannot explain ``why [the syllogisms of the proof] are linked together in one order rather than in another.'' 

To illustrate this gap more dramatically than the Euclid's proof example, consider \possessivecite{zagier1990one} one-sentence proof that every prime $p \equiv 1\pmod{4}$ is a sum of two squares (see \S\ref{app:zagier} for brief exposition of the proof). Verifying this proof requires only basic understanding of set properties, involution, and fixed-points. Each step can be checked for logical validity in a straightforward manner. However, such verification is not sufficient for understanding this proof (see \cite{31113} on how many mathematicians can easily verify \possessivecite{zagier1990one} proof but struggle to understand it). 

To see why, we consider some abilities required to demonstrate proof-understanding from \possessivecite{avigad2008understanding} framework (other criteria, for example those proposed by \citet{hamami2024understanding}, would similarly demonstrate the limitations). 

Consider the ability to ``indicate `key' or novel points in the argument, and separate them from the steps that are `straightforward'\,'': a world model can verify that the involution $f$ has exactly one fixed point, but it cannot identify why this step is the crucial insight versus the more routine verification that $f$ is indeed an involution. Similarly, for the ability to ```motivate' the proof, that is, to explain why certain steps are natural, or to be expected'': a world model cannot explain why constructing the specific set $S = \{(x,y,z) \in \mathbb{N}^3 : x^2 + 4yz = p\}$ was a natural choice, or why applying an involution to count fixed points would lead to the desired conclusion. Lastly, let us consider the ability to ``give a high-level outline, or overview of the proof'': while a world model can track the sequence of logical steps, it cannot provide the overarching strategy---that the proof works by cleverly counting the same set in two different ways using properties of involutions. 
Likewise, many other abilities posed in \cite{avigad2008understanding} present similar challenges to the world model conception of understanding.

\subsection{Understanding Physical Theories}
\label{subsec:understanding-physical-theories}

Understanding a physical theory requires more than simulating its internal mechanisms or predicting its outcomes---it involves understanding the problem situation that led to proposing that theory as a solution. This perspective, developed by \citet{popper1979objective}, defines ``problem situation'' as not only the problems one tries to solve but also the theoretical landscape that necessitated the solution---the inadequacies of existing theories and the specific explanatory gaps that generated new problems requiring solutions. 
In this perspective, understanding includes grasping the explanatory structure that motivated the theory's construction---why certain theoretical commitments were necessary to solve the identified problems and what makes this particular solution explanatorily adequate.

Consider Bohr's atomic theory \cite{bohr1913constitution}\footnote{For a refresher of Bohr's theory, see \S\ref{app:bohr}.}: Bohr proposed that electrons orbit the nucleus in discrete, fixed energy levels rather than in continuous paths as in classical physics. 
The key to understanding it is not merely visualizing, or world-modeling electrons jumping between orbits, but recognizing what Bohr attempted to explain with these electron jumps: the sharp, discrete spectral lines observable in atomic spectra.
To explain these definite, discrete lines, Bohr had to assume certain discreteness in electron movement possibilities, leading to the concept of jumps between tracks. 

Crucial to this explanation is the energy transfer mechanism Bohr proposed: when an electron jumps from an outer orbit to an inner orbit, the atom loses energy, which is emitted in the form of light radiation. The specific frequencies of light observed in spectral lines correspond directly to the energy differences between the allowed electron orbits. This mechanism explains why spectral lines appear at precise frequencies rather than a continuous spectrum.

Without knowing why Bohr introduced this somewhat unnatural model---to explain discrete spectral lines---one cannot understand his theory as a solution to a specific problem situation. The apparent unnaturalness of electrons being constrained to certain orbits and making quantum jumps between them only makes sense in light of the problem Bohr was solving. As \citet{popper1979objective} notes, someone who is just presented with the Bohr theory, without knowing that the theory was invented in order to explain the phenomenon of discrete spectral lines, will simply not understand the theory as a solution of a certain problem situation.

The world model alone (electrons orbiting in discrete paths) doesn't capture the understanding of the theory. World models in AI similarly emphasize internal simulation---like picturing electrons on orbits---but as this case-study shows, picturing is not understanding. An AI model might successfully simulate atomic transitions without grasping their importance in broader theoretical context, just as a human might visualize Bohr's orbital structure without understanding its explanatory role in solving the spectral line problem.

\section{A Possible Counterargument}
\label{sec:counterarguments}

A plausible counterargument can be proposed that the issue lies not with world models themselves, but with the level of abstraction we chose for states in our case studies. 
In our three case-studies, we selected states that align with prevailing conceptions in the literature---tracking objects like dominoes or electrons in the physical world. 
For mathematical reasoning, we likewise followed approaches like those of \cite{hu2023language}, which represent the prevailing conception in applying world models to this domain.
The counterargument suggests that world models could instead incorporate psychological or social abstractions as states themselves. For instance, in the domino computer example, the concept of primality and the statement ``641 is prime'' could be represented as states connected to the physical domino configurations. Similarly, for Bohr's theory, discrete spectral lines could be represented as states mapped to the mental picture of electrons orbiting in discrete paths.

While incorporating ad-hoc, abstract states connected with equally abstract transitions would rescue the world model approach from our critique, it would do so at the price of potentially undermining its falsifiability. If states can encode arbitrarily rich abstractions (mathematical properties, problem-solving strategies, historical context, explanatory motivations) then any phenomenon can be retrofitted into the world model approach simply by defining appropriate abstract states and transitions between them. The approach risks becoming irrefutable: any failure to capture understanding can be addressed by adding more complex states, and virtually any cognitive phenomenon can be accommodated by sufficiently enriching the state representations.

This flexibility limits the explanatory contribution of the world model concept itself. If the explanatory work is done by the state representations rather than the world model's dynamics, then the world model approach provides limited theoretical insight beyond organizing existing knowledge into states at various levels of abstraction and connecting them with equally abstract transitions. The counterargument essentially reduces to: ``world models can capture understanding if we put understanding into the states''---a circular explanation that presupposes what it seeks to explain.

\section{Conclusion}
\label{sec:conclusion}

In this paper, we argued that world models fall short of adequately characterizing human-level understanding. We supported this claim through case studies of particular, yet fundamental instances of understanding.
Although world models represent an advance over surface-level correlations, our case studies demonstrate important limitations in using world models as a lens for understanding. 
The world model research program represents, among current approaches, the most significant component, and the precursor to future theories of machine understanding. 
Our analysis offers only one perspective on how this research direction might be refined. 
The growing interest in world models as a path to artificial general intelligence makes it valuable to scrutinize whether this framework adequately captures what understanding entails. Philosophical examination of foundational concepts like understanding can complement algorithmic and experimental advances in addressing conceptual limitations of theoretical frameworks. While our perspective may be different from current trends in world model research, we believe such analysis contributes to ongoing discussions about the nature of understanding in artificial intelligence.

\appendix
\section{Hofstadter's Domino Chainium}
\label{app:domino}
\citet{hofstadter2007strange} introduces the thought experiment of a ``domino chainium''---a computer built from dominoes with special properties. In this system, each domino is spring-loaded so that after being knocked down, it automatically returns to its upright position after a short ``refractory'' period. This enables signals to propagate through the system repeatedly, supporting complex computational processes. The domino chainium functions as a mechanical computer where signals travel through carefully arranged networks of dominoes, bifurcating, joining, and propagating in loops to implement computer programs. The precise timing of domino falls determines how signals propagate and interact throughout the network.

In Hofstadter's example, this system is specifically designed to determine whether a number is prime. To test if a number is prime, one places exactly that many dominoes (e.g., 641) end-to-end in a designated ``input stretch'' of the network. When the first domino tips, it initiates a cascade that includes all the dominoes in the input stretch. This triggers a series of processes throughout the network, including various loops that test the input number for divisibility by different potential factors (2, 3, 5, etc.).

If any of these tests finds a divisor, a signal travels down a specific path called the ``divisor stretch,'' with falling dominoes indicating that the input number is not prime. Conversely, if all divisibility tests fail (meaning no divisors are found), a signal travels down a different path called the ``prime stretch,'' with falling dominoes confirming the number's primality. The system thus physically implements the algorithm for primality testing through the propagation of falling dominoes. The physical arrangement of dominoes embodies the logical structure of the primality test, with each part of the network serving a specific computational purpose---whether testing divisibility by a particular number, processing the results of these tests, or signaling the final outcome.

\section{Bohr's Atomic Theory}
\label{app:bohr}
Bohr's atomic theory \cite{bohr1913constitution}, proposed by Niels Bohr in 1913, was developed to address a specific problem in physics: explaining the discrete spectral lines emitted by atoms. When elements are heated or subjected to electrical discharges, they emit light that forms a unique pattern of discrete lines rather than a continuous spectrum when passed through a prism. This phenomenon contradicted classical physics, which predicted that electrons orbiting a nucleus would emit a continuous spectrum of electromagnetic radiation.

To explain these observations, Bohr introduced several radical postulates. First, he proposed that electrons can only orbit the nucleus in certain discrete, stable orbits (energy levels) where they do not emit radiation. Second, he suggested that electrons can jump between these allowed orbits. When an electron moves from a higher-energy orbit to a lower-energy orbit, it emits a photon with energy equal to the difference between the two orbital energy levels. The frequency of this photon corresponds directly to a specific spectral line.

This mechanism provided a direct explanation for why spectral lines appear at precise frequencies rather than forming a continuous spectrum. Bohr's model successfully explained the observed hydrogen spectrum and introduced the concept of quantization to atomic physics, laying groundwork for the development of quantum mechanics.

\section{Euclid's Proof of Infinite Primes}
\label{app:euclid}

Euclid's proof \cite{heath1956thirteen} that there are infinitely many prime numbers proceeds by contradiction. The proof begins by assuming that there are only finitely many prime numbers, which we can list as $p_1, p_2, p_3, \ldots, p_n$.
Given this assumption, Euclid constructs a new number $N$ by multiplying all these primes together and adding $1$:
$N = (p_1 \times p_2 \times p_3 \times \ldots \times p_n) + 1$

This number $N$ is now examined. There are two possibilities: either $N$ is prime, or $N$ is composite (not prime).
If $N$ is prime, then we have found a prime number not in our original list, contradicting our assumption that we had listed all prime numbers.
If $N$ is composite, then $N$ must be divisible by some prime number $q$. However, this prime $q$ cannot be any of the primes in our original list ($p_1, p_2, \ldots, p_n$) because dividing $N$ by any of these primes always leaves a remainder of $1$. Therefore, $q$ must be a prime number not in our original list, again contradicting our assumption.

Since both cases lead to a contradiction, our initial assumption must be false. Therefore, there must be infinitely many prime numbers.

\section{Zagier's One-Sentence Proof}
\label{app:zagier}

Zagier's original proof \cite{zagier1990one} that every prime $p \equiv 1 \pmod{4}$ is a sum of two squares is famously presented in a single sentence. Here we provide a slightly more detailed exposition of the argument.

\textbf{Theorem.} Every prime $p \equiv 1 \pmod{4}$ can be written as $p = a^2 + b^2$ for some integers $a, b$.

\textbf{Proof.} Consider the finite set $S = \{(x,y,z) \in \mathbb{N}^3 : x^2 + 4yz = p\}$. Since $x^2 \leq p - 4$ (as $y,z \geq 1$), we have $x \leq \sqrt{p-4}$, ensuring $S$ is finite. Define an involution $f: S \to S$ by:
\[
f(x,y,z) = \begin{cases}
(x + 2z, z, y - x - z) & \text{if } x < y - z \\
(2y - x, y, x - y + z) & \text{if } y - z < x < 2y \\
(x - 2y, x - y + z, y) & \text{if } x \geq 2y
\end{cases}
\]

This function is well-defined on $S$ and satisfies $f(f(x,y,z)) = (x,y,z)$ for all $(x,y,z) \in S$. The involution $f$ has exactly one fixed point: $(1, 1, (p-1)/4)$, where $p = 4k + 1$. By the \textit{parity principle}, if an involution on a finite set has an odd number of fixed points, then the set itself has odd cardinality. Since $f$ has exactly one fixed point, $|S|$ is odd.

Now consider the second involution $g: S \to S$ defined by $g(x,y,z) = (x,z,y)$, which simply swaps the $y$ and $z$ coordinates. Since $|S|$ is odd and $g$ is an involution, $g$ must have an odd number of fixed points by the parity principle. A fixed point of $g$ satisfies $(x,z,y) = (x,y,z)$, which means $y = z$. Thus there exists $(x,y,y) \in S$ such that $x^2 + 4y^2 = p$. Setting $a = x$ and $b = 2y$, we obtain $p = a^2 + b^2$, completing the proof.

\section{Perspectives on World Models and Understanding}
\label{app:quotes}

The following quotations illustrate a prevailing conception that links world models to understanding in artificial intelligence models.

\bigskip
\noindent\citet{lecun2022path}:
\begin{quotation}
    \textit{``Animals and humans exhibit learning abilities and understandings of the world that are far beyond the capabilities of current AI and machine learning (ML) systems. [...] By contrast, to be reliable, current ML systems need to be trained with very large numbers of trials [...]. Still, our best ML systems are still very far from matching human reliability in real-world tasks such as driving [...]. The answer may lie in the ability of humans and many animals to learn world models, internal models of how the world works.''}
\end{quotation}

\noindent\citet{ng2023aiunderstand}:
\begin{quotation}
    \textit{``Do large language models understand the world? [...] There's no widely agreed-upon, scientific test for whether a system really understands — as opposed to appearing to understand [...]. But with this caveat, I believe that LLMs build sufficiently complex models of the world that I feel comfortable saying that, to some extent, they do understand the world. To me, the work on Othello-GPT is a compelling demonstration that LLMs build world models; that is, they figure out what the world really is like rather than blindly parrot words.''}
\end{quotation}

\noindent\citet{mitchell2024worldmodels1}:
\begin{quotation}
    \textit{``Of course, the word ``understanding'' is ill-defined, but one thing that seems key to human understanding is having mental ``world models'': compressed, simulatable models of how aspects of the world work, ones that capture causal structure and can yield predictions.''}
\end{quotation}

\noindent\citet{delser2025worldmodelsartificialintelligence}:
\begin{quotation}
    \textit{``A crucial shortfall of modern AI is the lack of World Models. Unlike humans, who construct internal representations to reason, predict, and decide, large models rely solely on statistical associations in their training data. A World Model enables AI to develop a structured, dynamic understanding of its environment, capturing relationships, rules, and causal links.''}
\end{quotation}

\section*{Acknowledgements}
We thank the anonymous reviewers and Kinshuk Vasisht for their valuable feedback.
DP is grateful to Adobe Inc., Schmidt Sciences, Google and Microsoft Research for sponsoring his group's research.

\bibliography{aaai2026}

\end{document}